\documentclass[letterpaper, 10 pt, conference]{ieeeconf}  

\IEEEoverridecommandlockouts                              

\usepackage[noadjust]{cite}
\usepackage{makecell}

\usepackage{graphicx} 
\usepackage{subfigure}
\usepackage{amsmath} 
\usepackage{amssymb}  
\usepackage{multirow}
\usepackage{upgreek}
\usepackage{mathtools,lipsum, nccmath}
\usepackage{ulem}
\usepackage{xcolor}
\usepackage{caption}
\title{\LARGE \bf
Low-level Pose Control of Tilting Multirotor for Wall Perching Tasks Using Reinforcement Learning
}

\author{Hyungyu Lee$^{1}$, Myeongwoo Jeong$^{2}$, Chanyoung Kim$^{2}$, Hyungtae Lim$^{1}$, Changgue Park$^{3}$,\\ Sungwon Hwang$^{1}$, and Hyun Myung$^{1}$, \textit{Senior Member, IEEE}
\thanks{This research was supported by the National Research Foundation of Korea (NRF) Grant funded by the Ministry of Science and ICT for First-Mover Program for Accelerating Disruptive Technology Development (NRF-2018M3C1B9088328). The students are supported by the BK21 FOUR from the Ministry of Education (Republic of Korea).}

\thanks{$^{1}$Hyungyu Lee, $^{1}$Hyungtae Lim, $^{1}$Sungwon Hwang, and  $^{1}$Hyun Myung are with Urban Robotics Lab, KAIST}%
\thanks{$^{2}$Myeongwoo Jeong and $^{2}$Chanyoung Kim are undergraduate interns of Urban Robotics Lab, KAIST}%

\thanks{$^{3}$Changgue Park is with Korea Electronics Technology Institute(KETI)}}%

\begin{document}

\maketitle
\thispagestyle{empty}
\pagestyle{empty}

\begin{abstract}


Recently, needs for unmanned aerial vehicles (UAVs) that are attachable to the wall have been highlighted. 
As one of the ways to address the need, researches on various tilting multirotors that can increase maneuverability has been employed.
Unfortunately, existing studies on the tilting multirotors require considerable amounts of prior information on the complex dynamic model.
\textcolor{black}{Meanwhile, reinforcement learning on quadrotors has been studied to mitigate this issue.}
Yet, these are only been applied to standard quadrotors, whose systems are less complex than those of tilting multirotors.
In this paper, a novel reinforcement learning-based method is proposed to control a tilting multirotor on real-world applications, which is the first attempt to apply reinforcement learning to a tilting multirotor. 
To do so, we propose a novel reward function for a neural network model that takes power efficiency into account.
The model is initially trained over a simulated environment and then fine-tuned using real-world data in order to overcome the sim-to-real gap issue.  
Furthermore, a novel, efficient state representation with respect to the goal frame that helps the network learn optimal policy better is proposed.
As verified on real-world experiments, our proposed method shows robust controllability by overcoming the complex dynamics of tilting multirotors. 

\end{abstract}

\begin{keywords}
UAV, tiltable multirotor UAV, reinforcement learning
\end{keywords}

\vspace{-0.2cm}
\section{Introduction} \label{sec1}


The application of multirotors to replace various tasks in different fields has gained popularity in recent years \cite{choi, jung, caros1}. 
Despite the rapid development of multirotors, they are not yet suitable for tasks related to the maintenance of large structures such as skyscrapers or bridges that require external cleaning and inspection. That is because conventional quadcopters cannot maintain their position while changing the orientation due to the limited maneuverability.

    \begin{figure}[t] 
    \centering
        \includegraphics[scale=0.3]{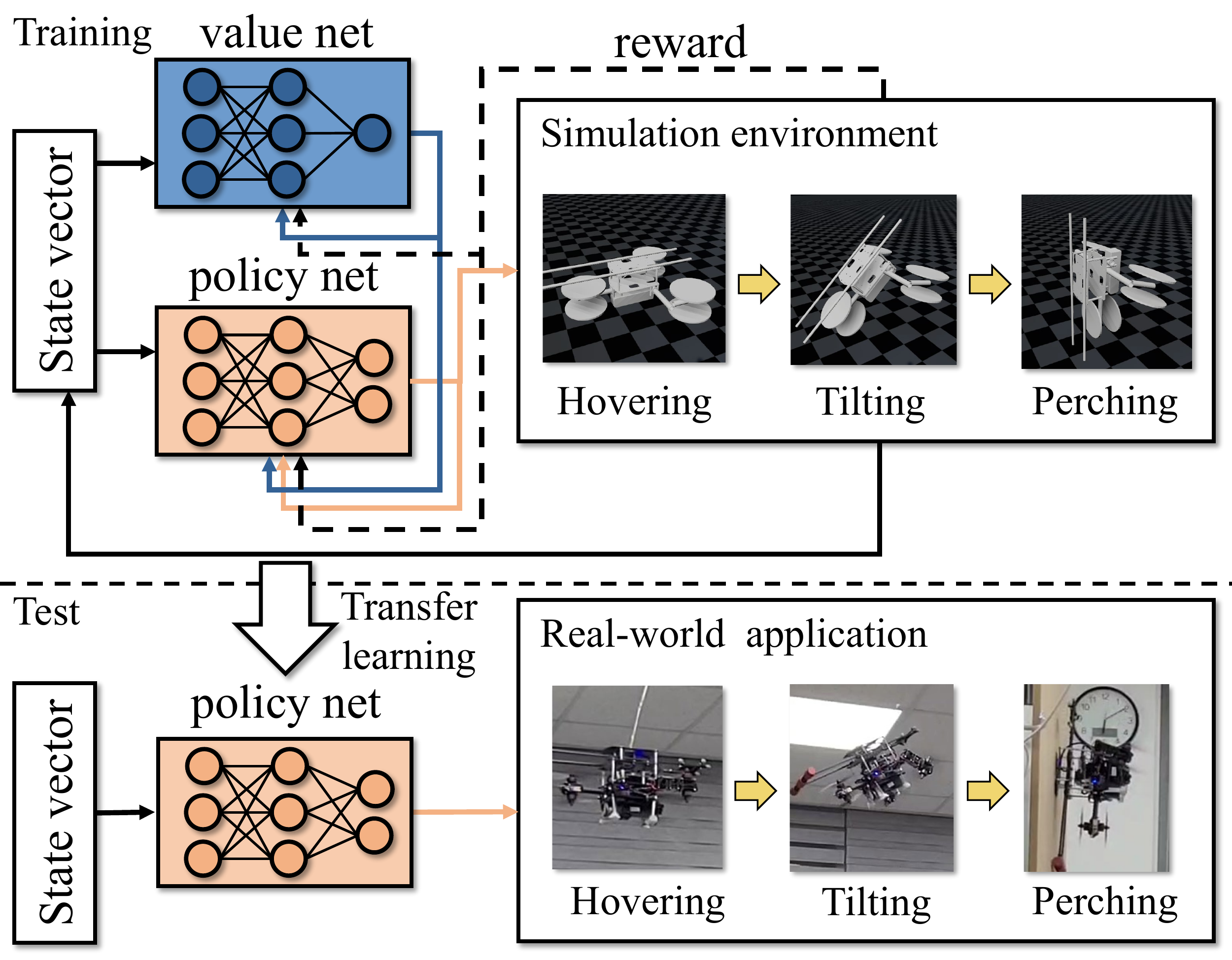}
        \caption{Schematic diagram of our entire process. In the simulation, the platform learns to hover, tilt, and perch sequentially using the policy and value networks. Through the learning with actual flight experiences, the nonlinear sim-to-real gap has been reduced. \textcolor{black}{This learning process using real-world data can be seen as transfer learning.}}
        \label{fig1}
            \vspace{-0.6cm}
    \end{figure}
\noindent

Various tilting multirotor UAVs have been studied to increase maneuverability \cite{caros, raj, bres, kam, oos, ryll, tog}. Some of the studies demonstrate a superior ability to respond to disturbances faster \cite{voy, seg} and perform complex tasks owing to improved mobility \cite{tog2}. 
However, because it is challenging to control the tilting multirotor owing to increased maneuverability, the control plays another critical role in tilting multirotor systems. Generally, multirotor control tasks are mainly categorized into two methodologies: a high-level control that performs mission, navigation, planning, and a low-level control that performs stabilization and motion control.
In this study, we place more emphasis on low-level control, particularly pose control. A multirotor is an unstable and nonlinear system by nature, which makes the control task to be challenging. Furthermore, in tilting multirotors, as servomotors are added to the multirotor system, the nonlinearity of the system increases due to interference caused by airflow between the tilting propellers. It is also more challenging to control due to the difference in response times between the servomotor and the rotor. To tackle these issues, various control strategies have been proposed. For instance, proportional-integral-derivative (PID) controllers are the renowned method \cite{caros,raj, bres, kam, oos, ryll, tog}, yet it requires accurate modeling and measurement of the system. Furthermore, many trials and errors are necessary to determine the optimal PID gain.


Meanwhile, various attempts have been made toward the control of multirotors at low levels using reinforcement learning to solve this classical control technique \cite{hwang,sergay,jjang }. These studies have broadened the scope of researches on controlling multirotors through reinforcement learning. A previous study used a controller learned in the simulation in an actual quadrotor to enable stable control under hovering and dynamic initial conditions \cite{hwang}. This approach shows that reinforcement learning trained in simulation environments opens the possibilities to low-level control real-world multirotors. However, the methodology does not directly address the sim-to-real gap issue.


To resolve the sim-to-real gap, a novel online learning method was proposed \cite{jjang} that uses real-world flight data to overcome the gap, showing the feasibility of learning real-world data on a quadrotor. Furthermore, another methodology attempt to learn directly on a real quadrotor using model-based reinforcement learning \cite{sergay}. However, the previous approaches are solely conducted on a quadrotor, where the sim-to-real gap is relatively smaller than that of a tilting multirotor. Furthermore, there has been a study to reduce the sim-to-real gap based on a previously trained module before the training on simulations called \textit{actuator net}, which helps the network learn the nonlinear relationship between input and output better \cite{anymal}. Unfortunately, it is a quadruped robot-based method, so it is not directly applicable on quadrotor platforms.


 



The higher complexity of the system due to additional actuators and the difference of response time between actuators are two main causes that make the tilting multirotor control a challenging problem.
We leverage the nonlinearity of the neural networks to tackle the issue, which can adapt to complex nonlinear approximation. Additionally, rotor outputs cannot be uniquely determined if the number of the actuators is larger than the force and moment expressed in the body frame. We have overcome these problems and present a novel control strategy that controls the tilting multirotor using reinforcement learning.

In summary, the contributions of this paper are fourfold:
\begin{itemize}
	\item To the best of our knowledge, it is the first attempt to perform low-level pose control of a tilting multirotor using reinforcement learning.
	\item We propose a goal-centric representation of pose that minimizes the amount of data required for optimal policy learning and reduces the ambiguity of the conventional state representation. Because the tilting multirotor is more complex than a conventional quadrotor system, more data is required for learning. The proposed pose representation method helps to overcome the problem.
	\item  An appropriate novel reward function is proposed by taking power consumption into account, which can be applied to platforms such as hexacopters, where the number of actuators is larger than the number of force and moment.
	The proposed reward function is more power-efficient and converges to a more stable policy compared to the conventional reward function used for the quadrotors. 
	\item The data augmentation method using the symmetry of drone dynamics suitable for the tilting multirotor domain was proposed. This can reduce the number of training data required for training.
\end{itemize}

The remainder of the paper is organized as follows. Section \uppercase\expandafter{\romannumeral2} introduces the background including the description of the platform. Section \uppercase\expandafter{\romannumeral3} describes the method, network structure, value network, and policy network learning method. Section  \uppercase\expandafter{\romannumeral4} presents the simulation environment, reward function, and the curriculum learning. The settings and results of the experiment are presented in Section \uppercase\expandafter{\romannumeral5} and conclusions are drawn in Section \uppercase\expandafter{\romannumeral6}.

\section{Background} \label{sec2}

\subsection{Tilting Multirotor Platform} \label{sec2a}

 \begin{figure}[t] 
   \centering
       \includegraphics[scale=0.45]{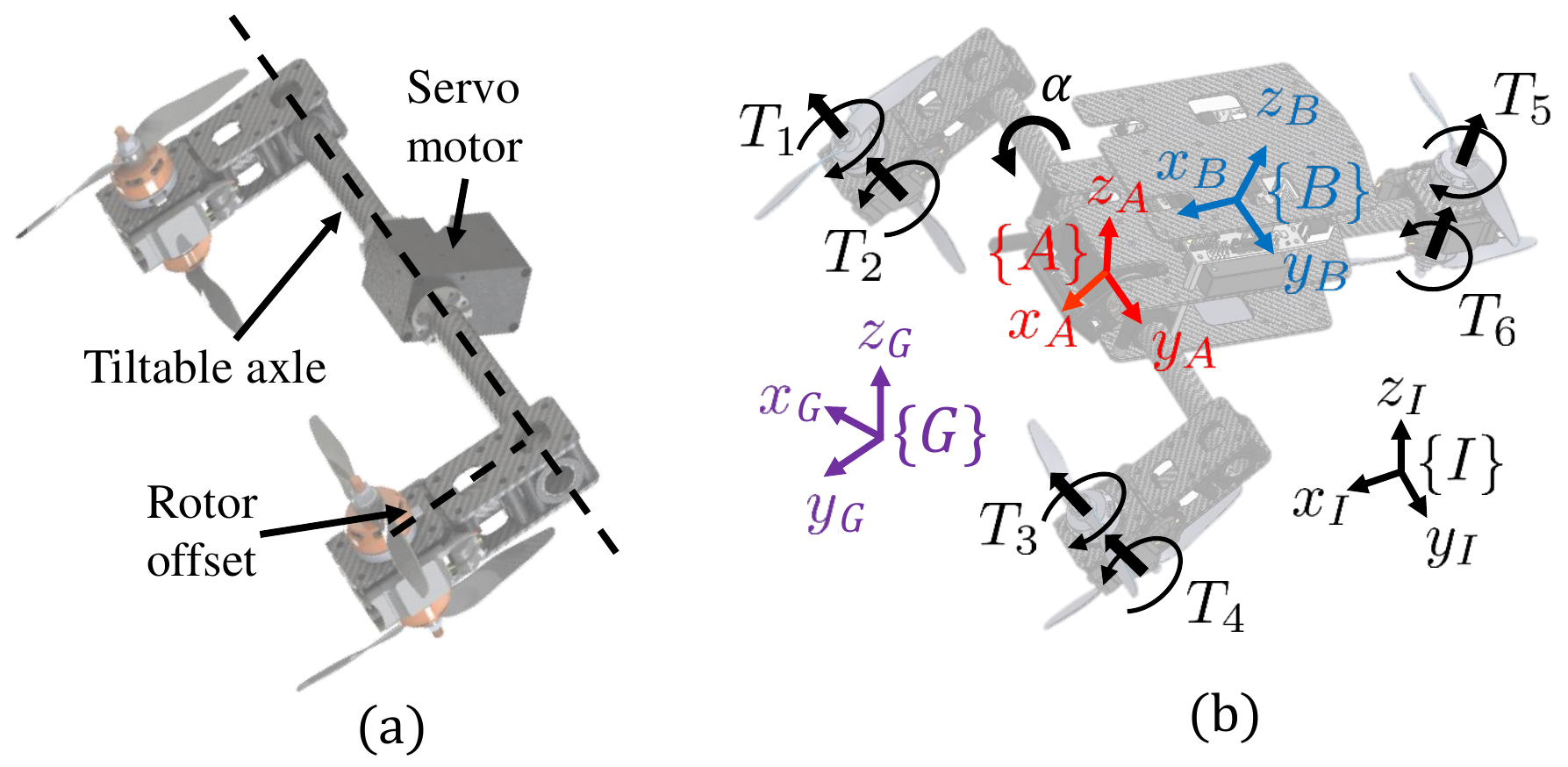}
       \caption{(a) The tilting mechanism of the platform. A servo motor rotates with respect to the tilable axle. (b) Coordinate  frames  of  the platform. $\{I\}$,$\{B\}$,$\{A\}$ and $\{G\}$ stand for the inertial, body, tilting axle, and goal frames, respectively.}        \label{fig2}  
       \vspace{-0.3cm}
    \end{figure}%

In this section, the platform and terminologies used in the study are summarized. The platform is based on a coaxial hexacopter, and in order for the platform to stably perch against a wall, it must be able to maintain its position even when the pitch of the drone is not zero. To make this physically possible, a servo motor is added to allow the propeller to rotate about the $y_{{A}}$-axis. 
To prevent the propeller from colliding to the wall during rotation, an offset is set between the rotor and the tilting shaft. Rotation angle of the servo motor angle $\alpha$ is constrained to range between 0\textdegree\ and 120\textdegree. Detailed effect of the rotor offset is not mentioned because it is beyond the scope of this paper.
$T_{i}$ represents the thrust of the $i$-th rotor. Body frame of the drone $\{B\}$ is located at the center of gravity. 
The arm frame $\{A\}$ is the frame rotated by $\mathrm{\alpha}$, and $T_1\mathrm{\ \sim \ }T_4$  have the same frame as $\{A\}$. $T_5\mathrm{\ \sim \ }T_6$ have the same frame as that of $\{B\}$. The goal frame $\{G\}$ is a frame rotated at $\psi$ in the $z$-axis with respect to the inertial frame. Assuming that the weight of the arm is negligible, the equation of motion can be expressed using the Newton-Euler equation for a single rigid body. The equation is as follows:
\begin{equation}
    \label{eq1}
         \begin{bmatrix} m\mathbf{I_3} & 0 \\ 0 & \mathbf{J}(\alpha) \end{bmatrix}
         \begin{bmatrix} \dot{\mathbf{v}}^B  \\  \dot{\boldsymbol{\upomega}}^B \end{bmatrix}
         +
         \begin{bmatrix} \boldsymbol{\upomega}^B \times (m \cdot \mathbf{v}^B) \\ \boldsymbol{\upomega}^B \times (\mathbf{J}(\alpha) \cdot \boldsymbol{\upomega}^B) \end{bmatrix}
         =
         \begin{bmatrix} \mathbf{F}^B \\ \mathbf{M}^B \end{bmatrix}
    \end{equation}
where $m$ is the mass, $\mathbf{J}(\alpha)$ inertial tensor with respect to $\mathbf{\alpha}$, $\mathbf{I_3}$ $3\times3$ identity matrix, $\mathbf{v}$ the linear velocity, $\boldsymbol{\upomega}$ the angular velocity, $\mathbf{F}=[F_x \; F_y \; F_z]^T$ the force, and $\mathbf{M}=[M_x \; M_y \; M_z]^T$ the moment. The superscript $B$ denotes the value represented in the body frame.

Additionally, the relationship between the frames can be established as follows:

\begin{equation}
\left[\begin{array}{l}
F_{x}^{B} \\
F_{z}^{B} \\
\textbf{M}^{B}
\end{array}\right]=\mathbf{A}(\alpha) \mathbf{T}
\end{equation}
where $\textbf{A}(\alpha)$ is a $5 \times 6$ allocation matrix which is a function of $\alpha$ and $\mathbf{T}=[T_{1}\; T_{2}\; T_{3}\; T_{4}\; T_{5}\; T_{6}]^T$ is the thrust vector.


\section{Method} \label{sec3}


\subsection{Network Structure} \label{sec3a}

\begin{figure}[t]
	\centering 
	\includegraphics[scale=0.23]{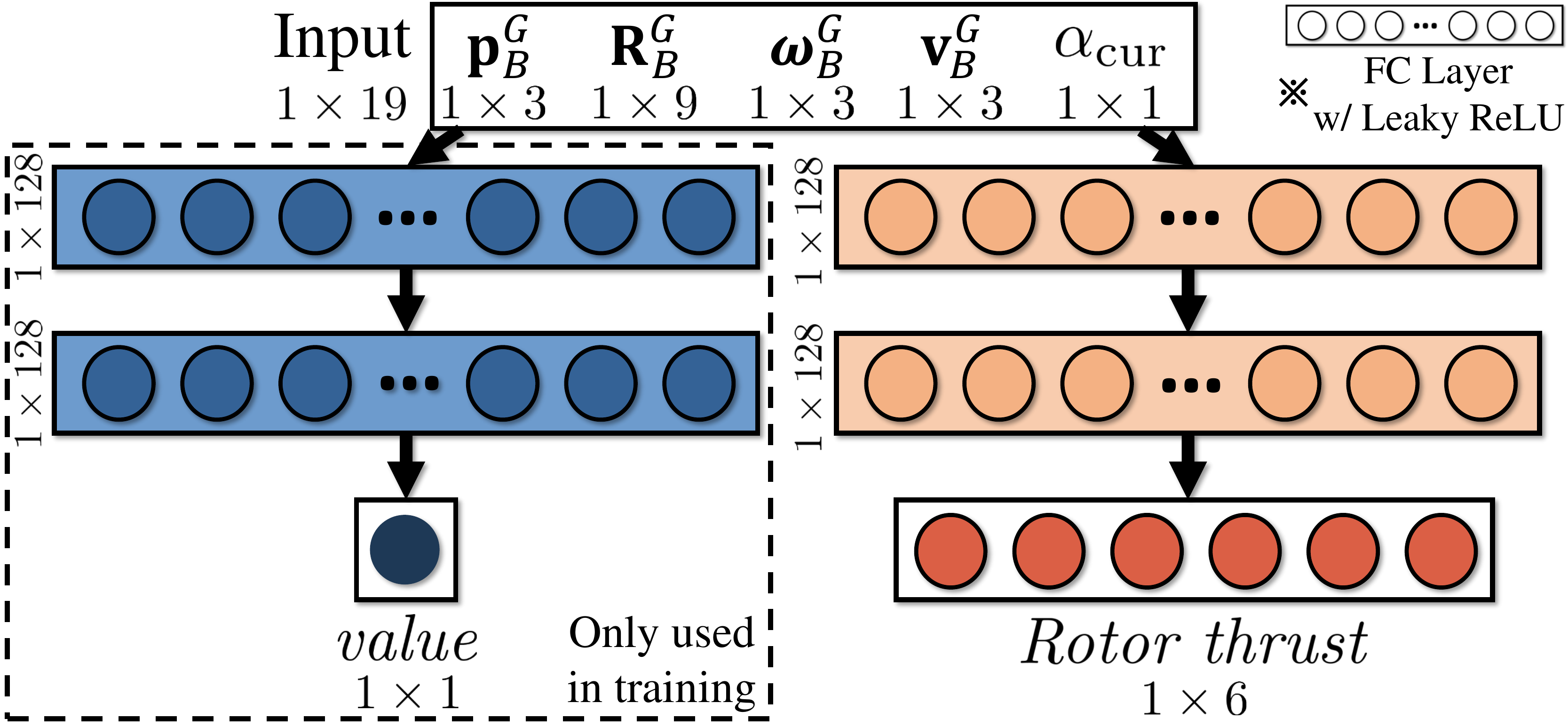}
	\caption{Overview of the proposed network that consists of value network and policy network. They both take $1\times19$ state as input and output value, and action, i.e. rotor thrust, respectively}
	\label{fig3}
	\vspace{-0.5cm}
\end{figure}

Because the network uses the Proximal Policy Optimization (PPO) algorithm based on the actor-critic method \cite{ppo}, our proposed neural networks consist of two parts: a) value network and b) policy network, which is shown in Fig.~\ref{fig3}. The networks receive 19-dimensional state vector as input respectively whose element consists of $\textbf{R}_{B}^{G}$, which is the relative rotation from the goal frame to the drone body frame, $\mathbf{{p}}_{{B}}^{G}$, $\mathbf{{v}}_{{B}}^{G}$, and $\boldsymbol{{\upomega}}_{{B}}^{G}$ the relative position, velocity, angular velocity of the drone expressed in the goal frame, and a current servo angle $\alpha_{\text {cur}}$. Each of the networks consists of two fully connected layers whose size is 128 with leaky ReLU. Finally, the value network outputs the value of the input state that is used to train the policy network that outputs $\mathbf{T}$. 


Even though the platform can follow the desired pitch $\theta_{\text {des}}$ while hovering, the reason for using the desired tilting angle $\alpha_{\text {des}}$ instead of $\theta_{\text {des}}$ is as follows. As a servo motor rotates along with $y_{A}$-axis, the corresponding unique $\theta$ value that makes the platform stay in the equilibrium point exists. If $\theta_{\text{des}}$ is used as an input, the neural networks must control the rotor and servomotor simultaneously. As a result, the real-to-sim gap increases due to the different response times between the servomotor and rotors. Besides, the dimension of the action space also increases, making it difficult to learn the optimal policy. Nevertheless, if $\alpha_{\text {cur}}$ is used by considering the characteristics of the platform whose hovering attitude is uniquely determined according to $\alpha_{\text {des}}$, the real-to-sim problem caused by the difference in the response time between the servo motor and rotors is eliminated, and the action space is reduced. We used an appropriate range of scaling such that the state follows a roughly normal distribution.

\subsection{Goal-centric Representation}

    \begin{figure}[t] 
    \centering
        \includegraphics[scale=0.23]{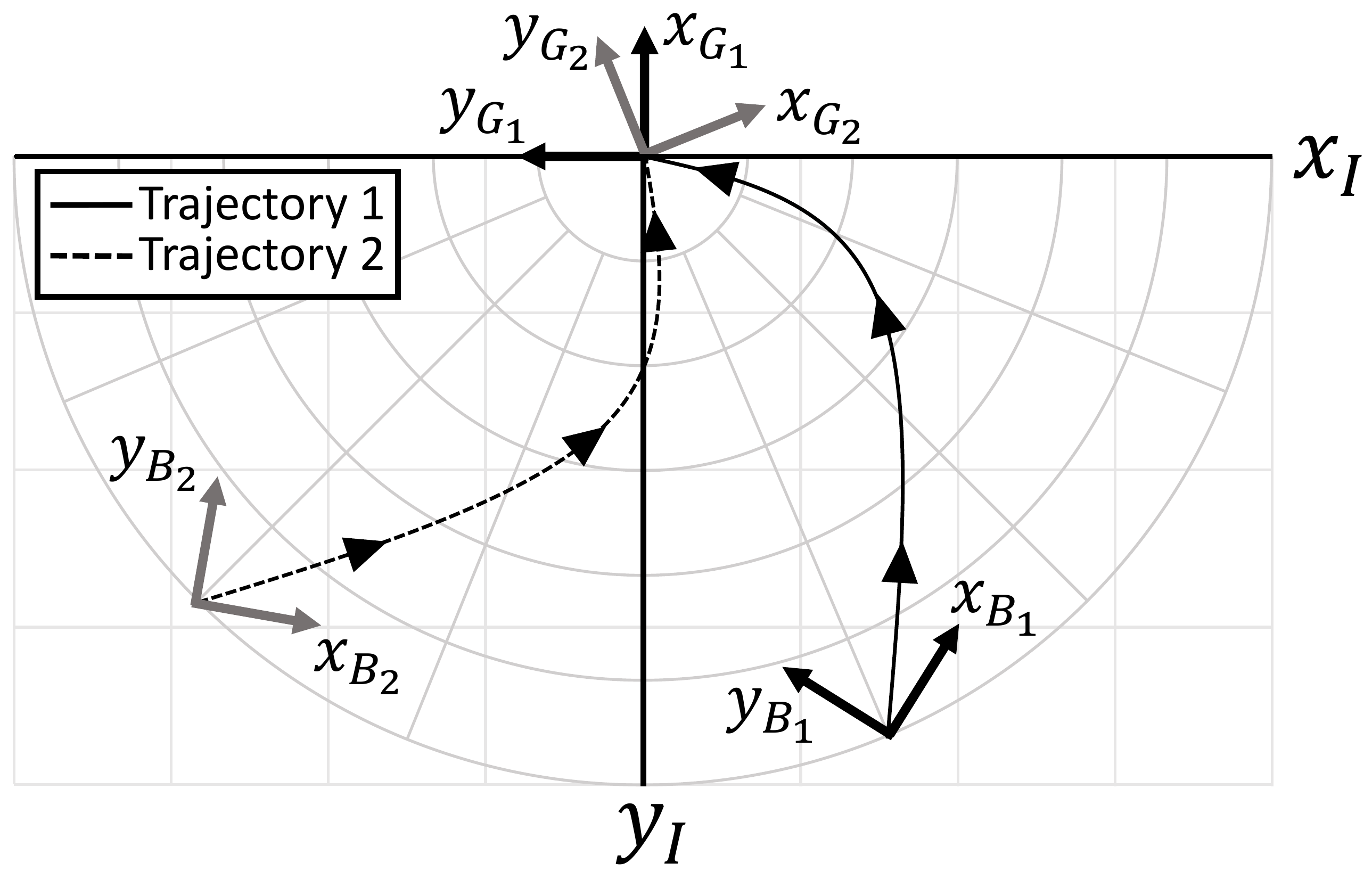}
        \caption{$xy$-plane view of the inertial frame. Trajectories 1 and 2 are starting from the equivalent state in goal-centric representation.}
        \label{fig4}
        \vspace{-0.3cm}
    \end{figure}

Note that our pose control is based on the goal frame, not the inertial frame. Intuitively, there will be a way to use the error between the drone's current pose and the goal pose based on a fixed inertial frame to control. In the conventional PID control, this method has no problem. However, when neural networks are used as the reinforcement learning controller, even if two initial poses follow the same goal-centric trajectory, the neural networks may recognize them as different states. In this case, the neural networks need an additional process to learn that the corresponding states make the same trajectory, slowing down the learning speed. A detailed description of this is as follows. The dynamic model of the drone's linear acceleration can be represented by the modeling above as follows:

\begin{equation}
\mathbf{R}_{B}^{I}\left(\mathbf{F}^{B}-\boldsymbol{\upomega}^{B} \times\left(m \mathbf{v}^{B}\right)\right)=m \left(\ddot{\mathbf{p}}^{I}-\mathbf{g}^{I}\right)
\end{equation}
where $\mathbf{g}^I$ is the gravity vector. Let $\mathbf{R}_{z^{I}}(\psi)$ be a rotation matrix that rotates by $\psi$ with respect to the $z$-axis $z^{I}$ of the inertial frame. Then, the gravity $\mathbf{g}^{I}$ has only z component, satisfying the following condition:

\begin{equation}
\mathbf{R}_{z^{I}}(\psi) \mathbf{g}^{I}=\mathbf{g}^{I}.
\end{equation}

Therefore, the following equation can be obtained using the above equations as:


\begin{multline}
\mathbf{R}_{z^{I}}(\psi) \mathbf{p}_{i}^{I}(t)=\mathbf{R}_{z^{I}}(\psi) \mathbf{p}_{i,0}^{I}+\mathbf{R}_{z^{I}}(\psi) \mathbf{v}_{i,0}^{I}t\ + \\\frac{1}{m}\left\{\mathbf{R}_{z^{I}}(\psi) \mathbf{R}_{B}^{I}(t)\int_{t_{0}}^{t} \int_{t_{0}}^{\tau^{\prime}}\Big(\mathbf{F}(\tau)^{B}-\boldsymbol{\upomega}^{B}(\tau)
\right.\\
\left.\times\left(m \cdot \mathbf{v}^{B}(\tau)\right) +\mathbf{g}^{I}\Big) \right. {d} \tau{d} \tau^{\prime} \Bigg\}
\end{multline}



\noindent
where $\mathbf{p}^{I}_{i}(t)$, $\mathbf{p}^{I}_{i, 0}$ and $\mathbf{v}^{I}_{i, 0}$ represent the position at time $t$, the initial position, and velocity of the ${i}$-th trajectory, respectively. The system dynamics is deterministic, and the policy network that outputs the system's only input $\mathbf{F}(\tau)^{B}$ is also deterministic when learning is complete. Therefore, when $\mathbf{p}_{1,0}^{I}=\mathbf{R}_{z^{I}}(\psi) \mathbf{p}_{2,0}^{I}, \mathbf{v}_{1,0}^{I}=\mathbf{R}_{z^{I}}(\psi) \mathbf{v}_{2,0}^{I}, \mathbf{R}_{B_{1}}^{I}(t)=\mathbf{R}_{z^{I}}(\psi) \mathbf{R}_{B_{2}}^{I}(t)$, and $\boldsymbol{\upomega}_{1,0}^{B}=\boldsymbol{\upomega}_{2,0}^{B}$ are satisfied, $\mathbf{p}_{1}^{I}(t)$ and $\mathbf{p}_{2}^{I}(t)$ must be equal. An example of this is shown in Fig.~\ref{fig4}. Trajectories 1 and 2 are starting from different initial positions $\mathbf{p}^I_{1,0}$ and $\mathbf{p}^I_{2,0}$ with zero initial linear/angular velocity, and flying toward goal frames $\left\{G_{1}\right\}$ and $\left\{G_{2}\right\}$. Because the condition of (5) is satisfied, the two trajectories have the same path when rotated about the $z^{I}$-axis. Therefore, the two initial states are essentially the same. However, there is a problem that the neural networks recognize it as a different state. This problem can be solved by expressing $\mathbf{p}_{B}^{I}, \mathbf{v}_{B}^{I}, \mathbf{R}_{B}^{I}$ vectors based on the goal frame reflecting $\mathbf{R}_{z^{I}}(\psi)$, thereby reducing the required data during learning.

\subsection{Data Augmentation}

    \begin{figure}[t] 
    \centering
        \includegraphics[scale=0.23]{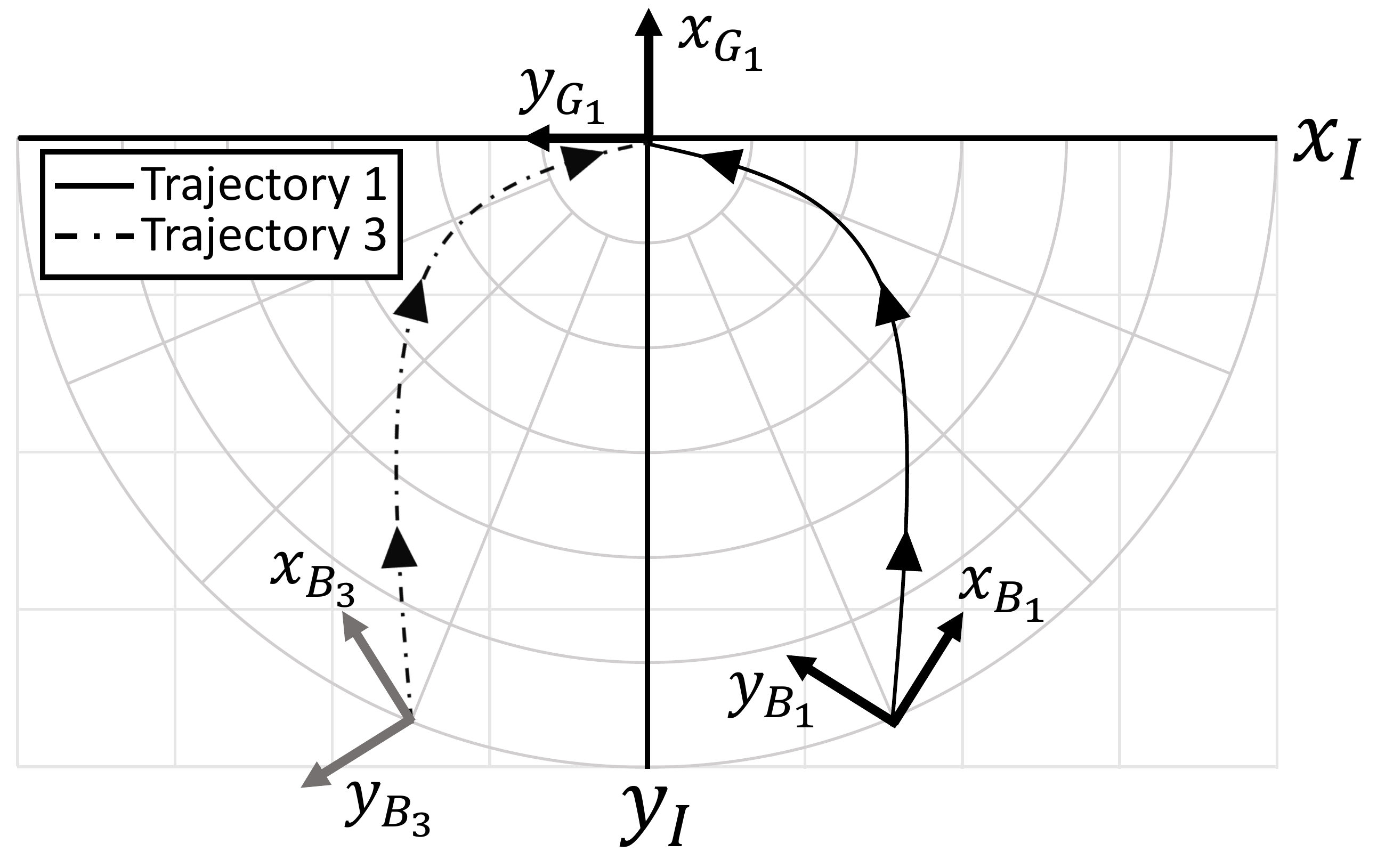}
        \caption{$xy$-plane view of the inertial frame. Trajectories 1 and 3 are starting from opposite states in goal-centric representation}
        \label{fig5}
        \vspace{-0.3cm}
    \end{figure}

In addition, data augmentation can be performed using the left-right symmetry of the drone. $\mathbf{T}^{\prime}$ represents the thrust vector when the platform's output thrust is geometrically inverted with respect to the $xz$-plane, as follows: $\mathbf{T}^{\prime}=\left(T_{3},T_{4}, T_{1}, T_{2}, T_{5}, T_{6}\right)^T$. Then, the following property is satisfied:

\begin{equation}
\left[\begin{array}{l}
F_{x}^{B} \\
F_{Z}^{B} \\
\overline{\mathbf{{M}}}^{B}
\end{array}\right]=\mathbf{A}(\alpha) \mathbf{T}^{\prime}
\end{equation}
where $\overline{\mathbf{M}}=\left({M}_{1},-{M}_{2}, {M}_{3}\right)$. Inertial to body rotation $\mathbf{R}_{B}^{I}(t)$ over time can be expressed as follows \cite{angular}:

\begin{equation}
{\boldsymbol{\mathrm{R}}}^I_B(t)=\mathrm{exp}\left[\frac{1}{2}\int^t_0{\ }\boldsymbol{\upomega}^{B}\left(t^{\mathrm{\prime}}\right){d}t^{\mathrm{\prime}}\right]{\boldsymbol{\mathrm{R}}}^I_B(0)
\end{equation}
where $\boldsymbol{\upomega}^{B}$ is defined as:

\begin{equation}
{\boldsymbol{\upomega }}^B(t)={\int^t_{t_0}{{\boldsymbol{\mathrm{J}}}^{{\mathrm{-1}}}\left(\alpha (\mathrm{\tau })\right)\boldsymbol{\mathrm{A}}{\left(\mathrm{\alpha }\mathrm{(}\mathrm{\tau }\mathrm{)}\right)}_{({3:5})}\mathbf{T}d\mathrm{\tau }}}.
\end{equation}

With these equations, the following equation holds:

\begin{multline}
\mathbf{p}_{i}^{G}(t)=\mathbf{p}_{i,0}^{G}+\mathbf{v}_{i,0}^{G} t \\+\frac{1}{m}\Bigg\{\exp \left[\frac{1}{2} \int_{t_{0}}^{t} \int_{t_{0}}^{\tau^{\prime}} \mathbf{J}^{-{1}}(\alpha(\tau)) \mathbf{A}(\alpha(\tau))_{(3:5)} \mathbf{T} {d} \tau {d} \tau^{\prime}\right]\\ \mathbf{R}_{B}^{G}(0) \int_{t_{0}}^{\tau^{\prime}} \int_{t_{0}}^{\tau}\left(\mathbf{A}(\alpha(\tau))_{(1: 2)} \mathbf{T}(\tau)\right.\\
\left.\left.-\boldsymbol{\upomega}^{B}(\tau) \times\left(m \cdot \mathbf{v}^{B}(\tau)\right.)+\mathbf{g}^{I}\right)\right. {d} \tau {d} \tau^{\prime}\Bigg\}.
\end{multline}

Simultaneously, the system and the optimal learned policy are deterministic with the same logic as before. Therefore ${\mathbf{{p}}}^{G_1}_{1,0}$ = ${\overline{\mathbf{p}}}^{G_1}_{2,0}$, $\mathbf{\ }{\mathbf{v}}^{G_1}_{1,0}$ = ${\overline{\mathbf{v}}}^{G_1}_{2,0}, \mathbf{\ }{\boldsymbol{\mathrm{R}}}^{G_1}_{{B_1}}(t)$ = ${\boldsymbol{\mathrm{R}}}_{z^{G_1}}(2\psi){\boldsymbol{\mathrm{R}}}^{G_1}_{{B_2}}(t),\ {\mathbf{p}}^{G_1}_{1,0}$ = ${\overline{\mathbf{p}}}^{G_1}_{2,0}$, and ${\boldsymbol{\upomega }}^{B_{\ }}_{1,0}$ = ${\overline{\boldsymbol{\upomega }}}^B_{2,0}$ are satisfied. From the relationship between  $\textbf{{A}}(\alpha )\mathbf{T}$ 
and $\textbf{A}(\alpha )\mathbf{T^{\prime}}$, the two trajectories are symmetric with respect to the plane passing through the origin that is perpendicular to ${{y}}_{{{G}}_{{1}}}$.

An example of this is shown in trajectories 1 and 3 in Fig.~\ref{fig5}. Trajectories 1 and 3 are starting from different initial positions $\mathbf{p}^{G_1}_{1,0}$ and $\mathbf{p}^{G_1}_{3,0}$ with  zero  initial linear/angular velocity, and flying toward the same goal frame $\left\{G_{1}\right\}$. 



\subsection{Value Network Training}
This section presents the learning of the value network. A value function $V_{\phi_{k}}$ that is parameterized with $\phi_{k}$ is updated using Monte-Carlo samplings obtained from a trajectory. This can be expressed by the following equation:

\begin{equation}
\phi_{k+1}=\underset{\phi}{\operatorname{argmin}} \frac{1}{K} \sum_{0}^{K}\left(V_{\phi}\left(s_{t}\right)-\hat{R}_{t}\right)^{2}
\end{equation}
where $K$ is the number of trajectories, ${\hat{R}}_t\ =\ \sum^T_{t^{\mathrm{'}}=t}{\ }R(s_{t^{\mathrm{'}}},$\\$a_{t^{\mathrm{'}}},s_{t^{\mathrm{'}}+1})$ reward to go, i.e., sum of rewards after the current position in a trajectory. Loss is obtained using least squares error. Through this predicted current value function $V_{\phi_{k}}$, the advantage estimate ${\hat{\mathcal{L}}}_t$ used in the actual policy optimization can be obtained. If an advantage estimate is made using well-tuned parameters, variance reduction is possible \cite{A2C} as follows:

\begin{equation}
\begin{split}
\hat{\mathcal{L}}_{t}&=\delta_{t}+(\gamma \lambda) \delta_{t+1}+\cdots+\cdots+(\gamma \lambda)^{K-t+1} \delta_{T-1}\\
\delta_{t}&=r_{t}+\gamma V\left(s_{t+1}\right)-V\left(s_{t}\right)
\end{split}
\end{equation}
where $r_t$ is the reward at step $t$ and $\lambda\ \mathrm{\in }\ \mathrm{[0,1]}$ a hyper parameter.

\subsection{ Policy Network Training}

\begin{center}
\small{TABLE \uppercase\expandafter{\romannumeral1}

\textsc{Parameters Used For Training In The Simulation}}
\end{center}
$$
\small{}
\begin{array}{lc}
\\[-0.5ex] 
\hline
\hline
\\[-1ex] 
\multicolumn{1}{c} { \text{Parameter} } &  { \text{Value} } \\[1ex] 
\hline \\[-1ex]
\text { Number of steps per environment } & 1,200 \\
\text { Epoch size } & 10 \\
\text { Batch size } & 10 \\
\text { Value function coefficient }\left(\mathrm{c}_{1}\right) & 0.5 \\
\text { Learning rate } & 0.00005 \\
\text { Discount factor }(\gamma) & 0.998 \\
\text { GAE factor }(\lambda) & 0.95 \\
\text { Clipping parameter }(\epsilon) & 0.2 \\
\text { Optimizer } & \text { Adam }\\
\\[-1ex] 
\hline
\hline
\end{array}
$$

\normalsize{}

The policy optimization process is calculated using ${\mathcal{L}}_t$ obtained in equation (11). The loss function used at this time is PPO-clip presented in \cite{ppo}. A parameter $\theta_{k}$ for policy function $\pi_{\theta_{k}}$ is updated as follows:
\begin{equation}
\begin{array}{l}
\theta_{k+1}=\underset{\theta}{\operatorname{argmax}} \frac{1}{K} \sum_{t=0}^{K} \bigg\{\min \Big(\frac{\pi_{\theta}\left(a_{t} \mid s_{t}\right)}{\pi_{\theta_{k}}\left(a_{t} \mid s_{t}\right)} \mathcal{L}_{t}^{\pi_{\theta_{k}}}, \\
\quad\quad\quad\quad\quad g\left(\epsilon, \mathcal{L}_{t}^{\pi_{\theta_{k}}}\right)\Big)-c_{1}\left(V_{\phi}\left(s_{t}\right)-\widehat{R}_{t}\right)^{2}\bigg\}
\end{array}
\end{equation} 
\noindent
where $g(\epsilon, \mathcal{L})=\left\{\begin{array}{ll}(1+\epsilon) \mathcal{L}, & \mathcal{L} \geq 0 \\ (1-\epsilon) \mathcal{L}, & \mathcal{L}<0\end{array}\right.$\\ and $\mathcal{L}_{t}^{\pi_{\theta_{k}}}$ is $\hat{\mathcal{L}_{t}}$ under policy $\pi_{\theta_{k}}$; ${\mathrm{c}}_{\mathrm{1}}$ and $\epsilon$ are hyper parameters that satisfy ${\mathrm{c}}_{\mathrm{1}}, \epsilon \ \mathrm{\in }\ \mathrm{[0,1]}$.

\section{Simulation}
This section presents the learning process through simulations, including the simulator used, simulation parameters, reward function, and the simulation results.

\subsection{ Simulation Environment }

For the simulator, we used Raisim \cite{raisim}, which guarantees faster speed and accuracy than other simulators, and its performance was verified in various previous studies \cite{hwang, anymal}. Parameters used in the simulation are listed in Table \uppercase\expandafter{\romannumeral1}. To learn various samples, values within the ranges of $\mathrm{\pm}$1 m, $\mathrm{\pm}$1 m/s, and $\mathrm{\pm}$1 rad/s are randomly assigned to the position, linear velocity, and angular velocity, respectively. The initial orientation is also given randomly within the range where the body $z$-axis has a positive value. The training was performed on a Linux machine with an Intel Core i7-6700K CPU @ 4.00GHz, and GPU GTX 1660.

\subsection{Reward Function}

    \begin{figure}[t] 
    \centering
        \includegraphics[scale=0.42]{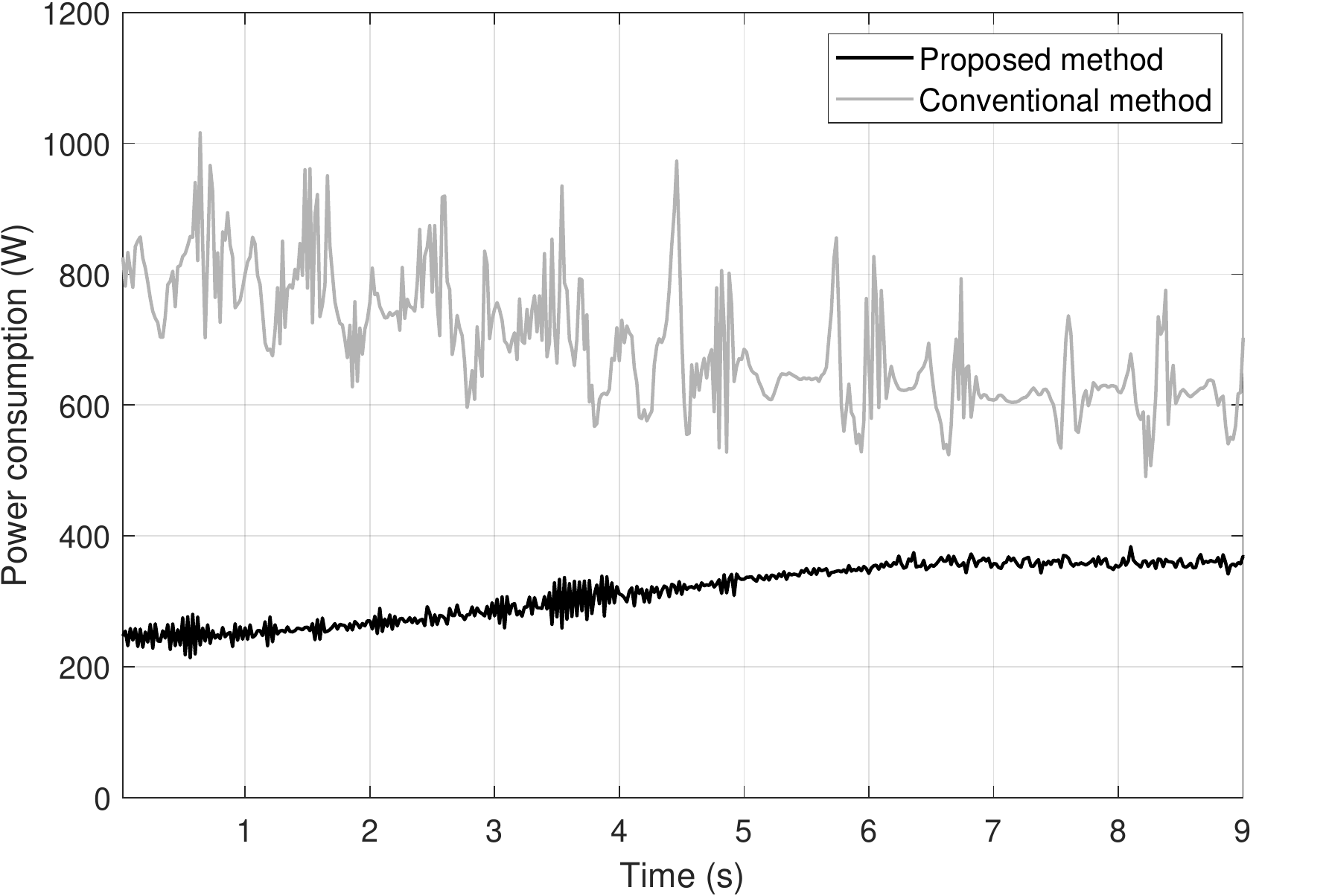}
        \caption{ Power consumption at each timestep while hovering with arm tilting angle from 0\textdegree \ to 110\textdegree.}
        \label{energy}
        \vspace{-0.4cm}
    \end{figure}

To control the system to achieve the desired performance through reinforcement learning, an appropriate cost function must be set. The cost function used is as follows:
\begin{equation}
r_{t}= c_p\left\|\mathbf{p}_{t}\right\|+c_\psi\left\|{\psi}_{t}\right\|+c_{\boldsymbol{\upomega}}\left\|\boldsymbol{\upomega}_{t}\right\|+c_E\left\|{E}_{t}\right\|
\end{equation}
where ${\mathbf{p}}_t$, ${\psi}_t,\ {\boldsymbol{\upomega }}_t$, and ${E}_t$ are the position, angle with respect to $z^{G}$-axis, angular velocity, and power consumption at time \textit{t}, respectively. $c_{p}$, $c_{\psi}$, $c_{\boldsymbol{\upomega}}$, and $c_E$ are the weighting factors for ${\mathbf{p}}_t$, ${\psi}_t,\ {\boldsymbol{\upomega }}_t$, and ${E}_t$, respectively. In the case of attitude reward, our platform does not use $\theta_{des}$, but uses $\alpha_{des}$, and learns the optimal attitude for the current tilting angle $\alpha_{cur}$ in the learning process.
Therefore, the angle difference between the desired and the current $\psi$ was used as an error.
The quadrotor, which was mainly used for low-level multirotor control through reinforcement learning, has a system whose four actuators output $F_{z}^B, M_{x}^B, M_{y}^B,$ and $M_{z}^B$. However, because our platform is based on a coaxial hexacopter, seven actuators are redundant, outputting only $F_{x}^B, F_{z}^B, M_{x}^B, M_{y}^B,$ and $M_{z}^B$. Therefore, there are various solutions that make the actuator generate the equivalent outputs. Because the policy network needs to converge to an optimal policy, such a situation is undesirable. Thus, power consumption was considered as a reward function, which was not considered in the previous reinforcement learning studies using quadrotors. The relationship between thrust and power consumption in the idle rotor system can be expressed as follows \cite{gem}:
$$
P=T \cdot \sqrt{\frac{T}{2 A \rho}}
$$
\noindent 
where $A$ is the rotor disk area, $\rho$ is the density of the air, $P$ the required power for the thrust $T$. The comparison result between the proposed and conventional methods is shown in Fig. \ref{energy}. Because the reward function used in the previous studies was focused on a quadrotor, it was not necessary to consider the power consumption from thrust. However, in our platform, the power consumption itself is more significant than a quadrotor, and undesirable oscillation occurs in the thrust. Because the rotor output is also nonlinear, such a thrust output should be avoided to prevent a more significant sim-to-real gap. On the other hand, when the proposed method is used, the results are much more stable than those with the conventional method. For efficient learning, the platform converges to the origin of the inertial frame at the time of learning. In the actual situation, rewards are expressed in the goal frame, which is given by the manual input. Because the position and orientation are the parts we care about most, it has the highest coefficient.

\subsection{Curriculum Learning }

Unlike the conventional quadrotor, the tilting multirotor has more complex dynamics and systems. Therefore, it takes a long time to learn the optimal policy, or it can converge to the local optima. Complex problems can be more effectively solved using curriculum learning \cite{curri,curri2}. These studies show that learning through curriculum learning using a more accessible example promotes the neural networks to converge faster. The curriculum was conducted by learning to hover while reducing the randomness of the initial condition. Then, the randomness is gradually increased, after which the tilting and perching are learned while increasing the tilting angle. Thus, the agent can converge faster and overcome the problem of falling into a local optimum during learning. 

Also, domain randomization has been used with curriculum learning. Domain randomization is a concept suggested in \cite{dom} to reduce the real-to-sim gap in reinforcement learning. It creates a simulation environment with random characteristics such as wind, and random noise is applied to sensors and hardware while learning. Wind disturbances, sensor noises, and hardware parameters are randomized so that the initial optimal policy can run in reality. However, it acts as a factor that makes learning to be complicated. So curriculum learning was performed by applying 5\% noise to sensor data and hardware parameters.

\subsection{Simulation Result}
    \begin{figure}[t] 
    \centering
        \includegraphics[scale=0.57]{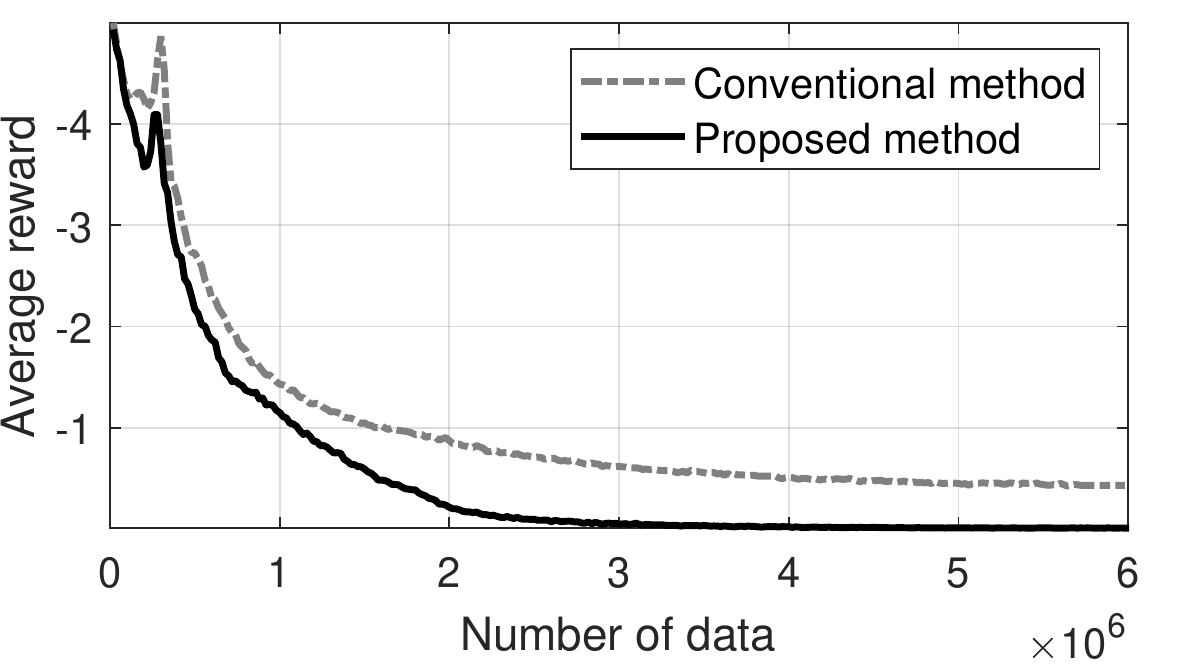}
        \caption{Learning curves when hovering is trained using the conventional and proposed method. \textcolor{black}{The number of data is equal to the total number of steps.}}
        \label{sim_cost}
        \vspace{-0.2cm}
    \end{figure}
        \begin{figure}[t] 
    \centering
        \includegraphics[scale=0.42]{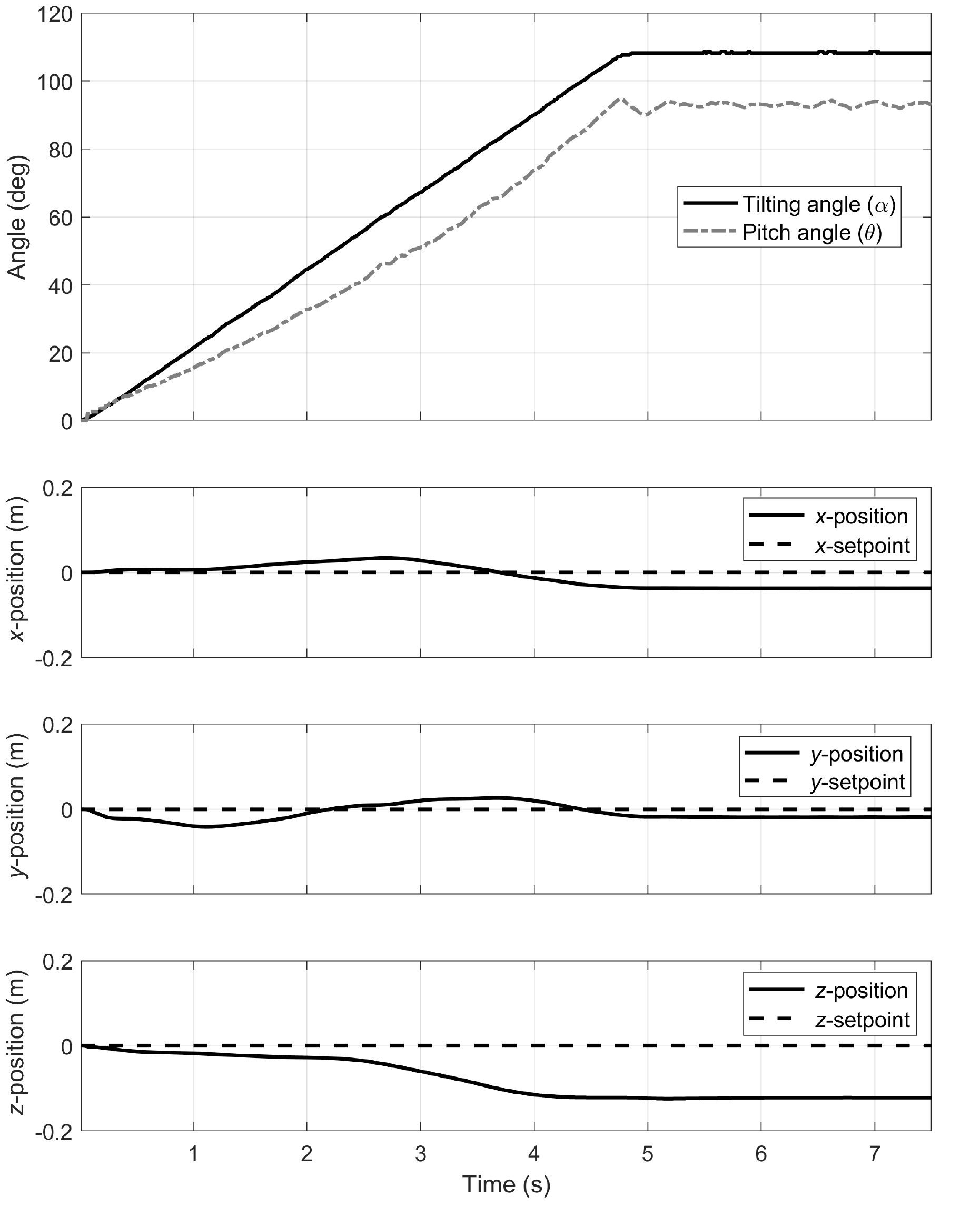}
        \caption{\textcolor{black}{Simulation results using learned optimal policy while hovering.} The pitch angle $\theta$ changes according to the given tiling angle $\alpha$.}
        \label{sim_log}
        \vspace{-0.5cm}
    \end{figure}
            \begin{figure}[t] 
    \centering
        \includegraphics[scale=0.45]{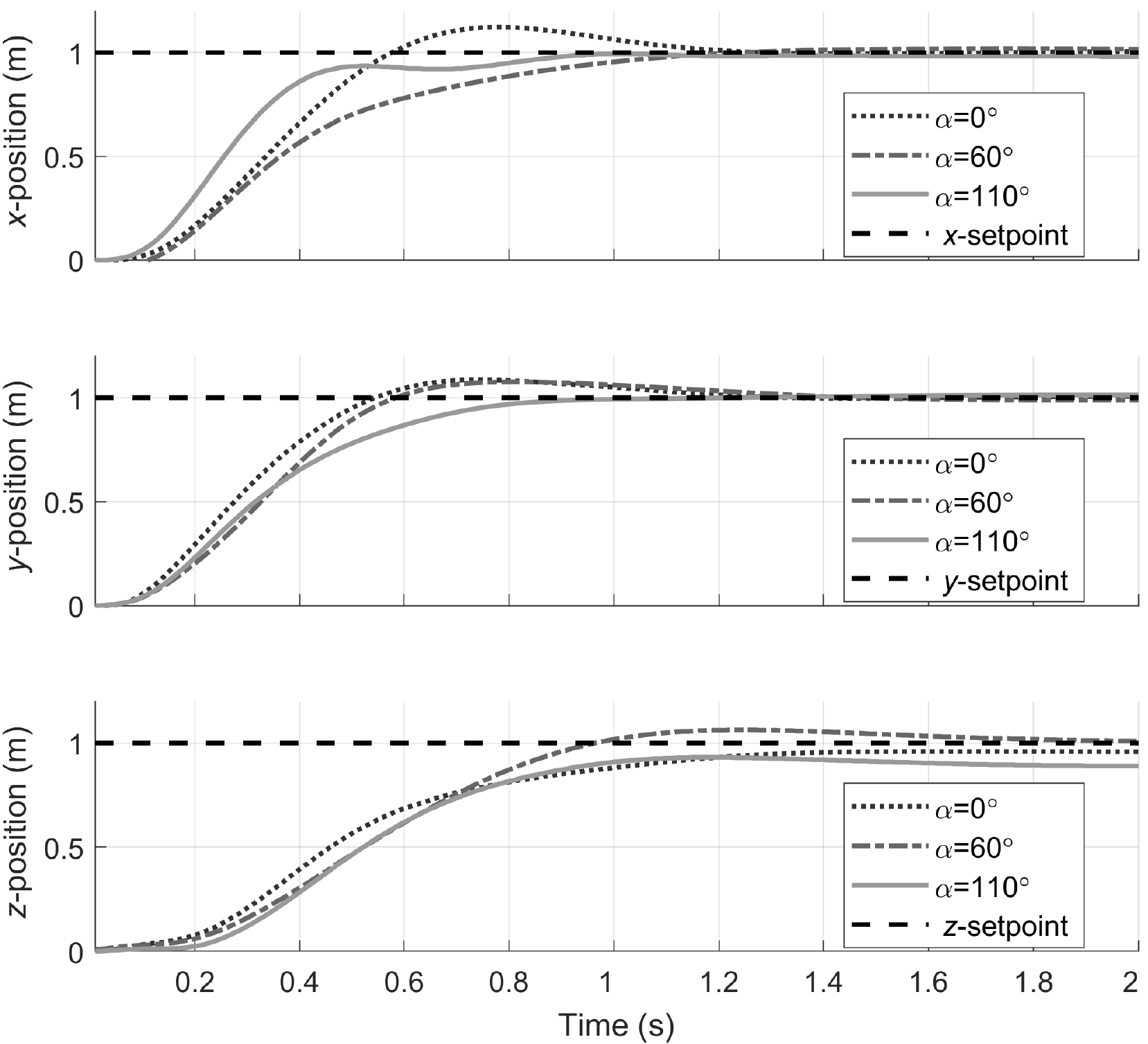}
        \caption{\textcolor{black}{Position step responses of the learned optimal policy in the simulation at various tilting angles.}}
        \label{sim_step_log}
        \vspace{-0.2cm}
    \end{figure}

    The convergence speed when learning hovering in the simulation is checked to determine the extent to which the contribution affects the result. As shown in Fig.~\ref{sim_cost}, the network was thoroughly trained with 3 million data using the proposed method, while the conventional method did not converge. Consequently, it can be confirmed that the proposed method reduces the data required for learning, resulting in faster convergence.
    Two tests were conducted to confirm the performance of the policy learned in the simulation. It is necessary to decouple the $x$-position and the angle $\theta$ to perch on the wall stably. 
    \textcolor{black}{For the first test, we checked if the learned optimal policy could maintain the desired position with an adequate pitch angle $\theta$ in mid-air while the tilting angle $\alpha$ changed from 0\textdegree to 110\textdegree.} The results are shown in Fig.~\ref{sim_log}. Even if $\alpha$ changes, the platform maintains its position with the appropriate $\theta$. However, it is noteworthy that the $z$-position gradually decreases as $\alpha$ increases. It is because the efficiency of the platform decreases as the arm is tilted, and the power consumption by thrust is considered in the reward function. Additionally, to check the control performance according to $\alpha$, the test using a step position input was carried out. The results are shown in Fig.~\ref{sim_step_log}. $\alpha$ of 0\textdegree, 60\textdegree, 110\textdegree\space are the tilting angles that make the platform hover at 0\textdegree, 45\textdegree, 90\textdegree, respectively. The behavior of the platform differs depending on ${\alpha}_{cur}$; however, it was clear that it converged to the setpoint. For the $z$-position, it can be seen that the steady-state error increases as $\alpha$ increases, similar to the result in Fig.~\ref{sim_log}. From these results, we can conclude that the policy was sufficiently learned to perform the actual flight.

\section{Experiments}
This section describes the hardware parameters, overall operation diagram of the platform used in the experiment, and experimental results. For the experiment, the flight performance of the pose control between the cases with and without learning the actual experience was compared.

\subsection{Hardware}
    \begin{figure}[t] 
    \centering
        \includegraphics[scale=0.63]{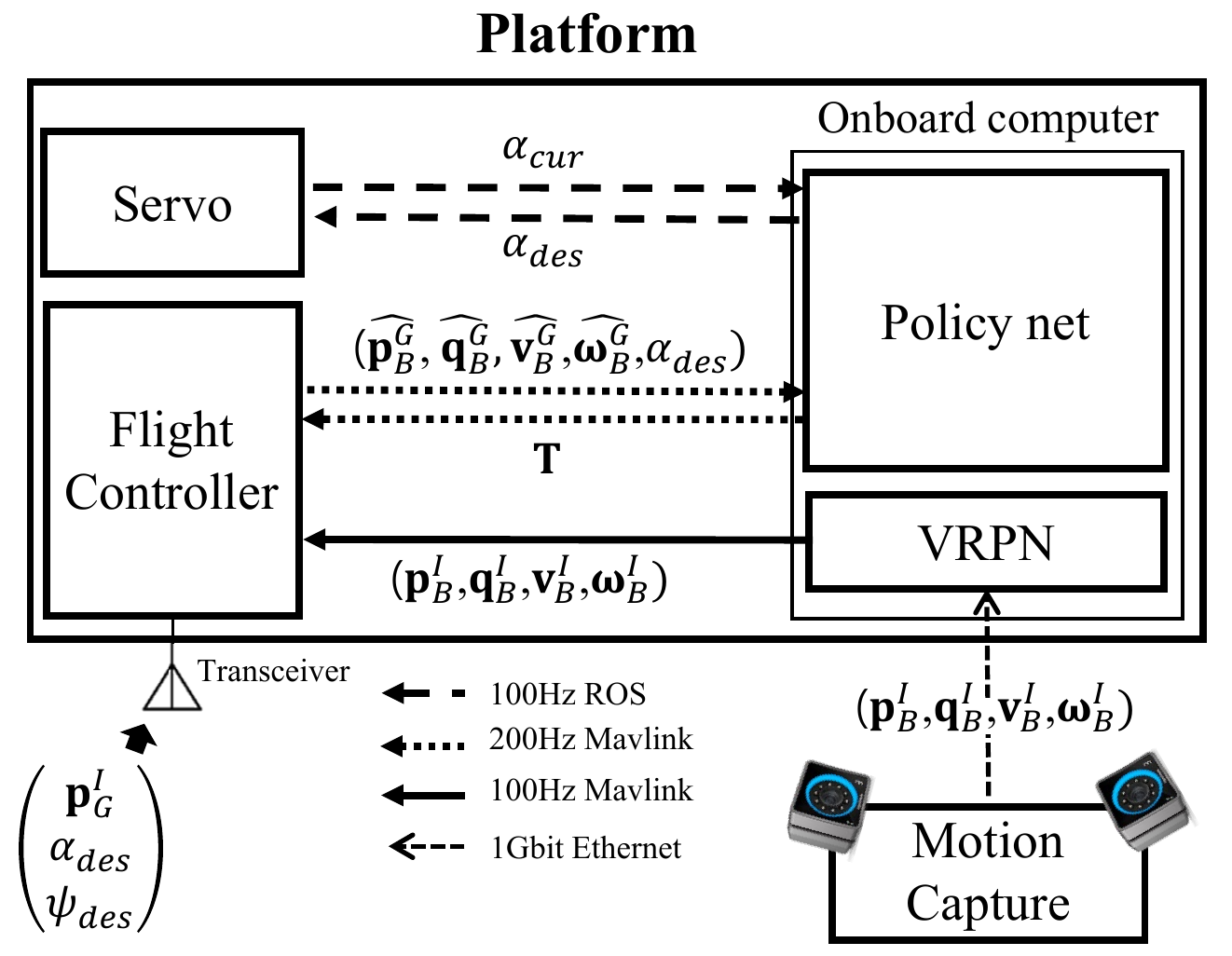}
        \caption{Platform diagram for experimentation. The 1Gbit Ethernet uses wireless communication, and the Mavlink and ROS use wired communication.}
        \label{platform}
        \vspace{-0.4cm}
    \end{figure}

The main components of the system consist of 5-inch propellers, Xnova RM2207 2000 KV motors \cite{tmot}, and Robotis Dynamixel XM540-W270-R \cite{rob}. Pixhawk4 mini \cite{pix} was used as a flight controller, and Jetson Nano with Quad-core ARM Cortex-A57 MPCore processor is used for the onboard controller. The main parameters are listed in Table \uppercase\expandafter{\romannumeral2}.

\small{}
\begin{center}
\small{TABLE \uppercase\expandafter{\romannumeral2}

\textsc{Hardware Parameters}}
\end{center}
$$
\small{}
\begin{array}{lc}
\\[-1ex] 
\hline
\hline
\\[-1ex] 
\multicolumn{1}{c} { \text{Parameter} } &  { \text{Value} } \\[1ex] 
\hline 
\\[-1ex]
\text { Total weight } & 1.9 \mathrm{~kg} \\
\text { Max. thrust of each BLDC motor \cite{tmot} } & 14.2 \mathrm{~N} \\
\text { Stall torque of XM540-W270-R \cite{rob} } & 10.60 \mathrm{Nm} \\
\text { Distance from tilting axle to tilting rotor} & 75 \mathrm{~mm}\\
\text { Half of the distance between tilting rotors} & 130 \mathrm{~mm}  \\
\text { Distance from tilting axle to fixed rotor } & 200 \mathrm{~mm}\\
\\[-1ex] 
\hline
\hline
\label{tab2}
\end{array}
$$

\normalsize{}

A diagram of the entire system is detailed in Fig.~\ref{platform}, where $\hat{\mathbf{x}}$ represents the estimation of the vector $\mathbf{x}$. After ${\mathbf{p}}^{{I}}_{B}$, ${\mathbf{q}}^{{I}}_{B}$, ${\mathbf{v}}^{{I}}_{{B}}$, and ${\boldsymbol{\upomega }}^{{I}}_{{B}}$ are measured by the motion capture, they are sent to the VRPN node of the on-board computer running the ROS (Robot Operating System). This process was performed through wireless communication. The VRPN node sends the data to the flight controller using a protocol referred to as Mavlink. Here, the user inputs ${\mathbf{p}}^{{I}}_{G}$, ${\alpha}_{des}$, and ${\psi}_{des}$ to the flight controller through the radio controller. Then, estimated states are transmitted to the policy network node through Mavlink. In the policy network node, state vectors are inserted into the learned policy network to inference \textbf{{T}}. After that, \textbf{{T}} and ${{\alpha }}_{{des}}$ are inputted to the flight controller and servo motor through Mavlink and ROS protocols, respectively. In the flight controller, the input \textbf{{T}} is converted into PWM and transmitted to each rotor. In the servo motor, the PID control is performed through the input ${{\alpha }}_{{des}}$. Subsequently, the tasks described above are repeated as the policy network node gets state feedback.
\subsection{Without \textcolor{black}{Real-world Data}}
\begin{figure}[t] 
    \centering
        \includegraphics[scale=0.42]{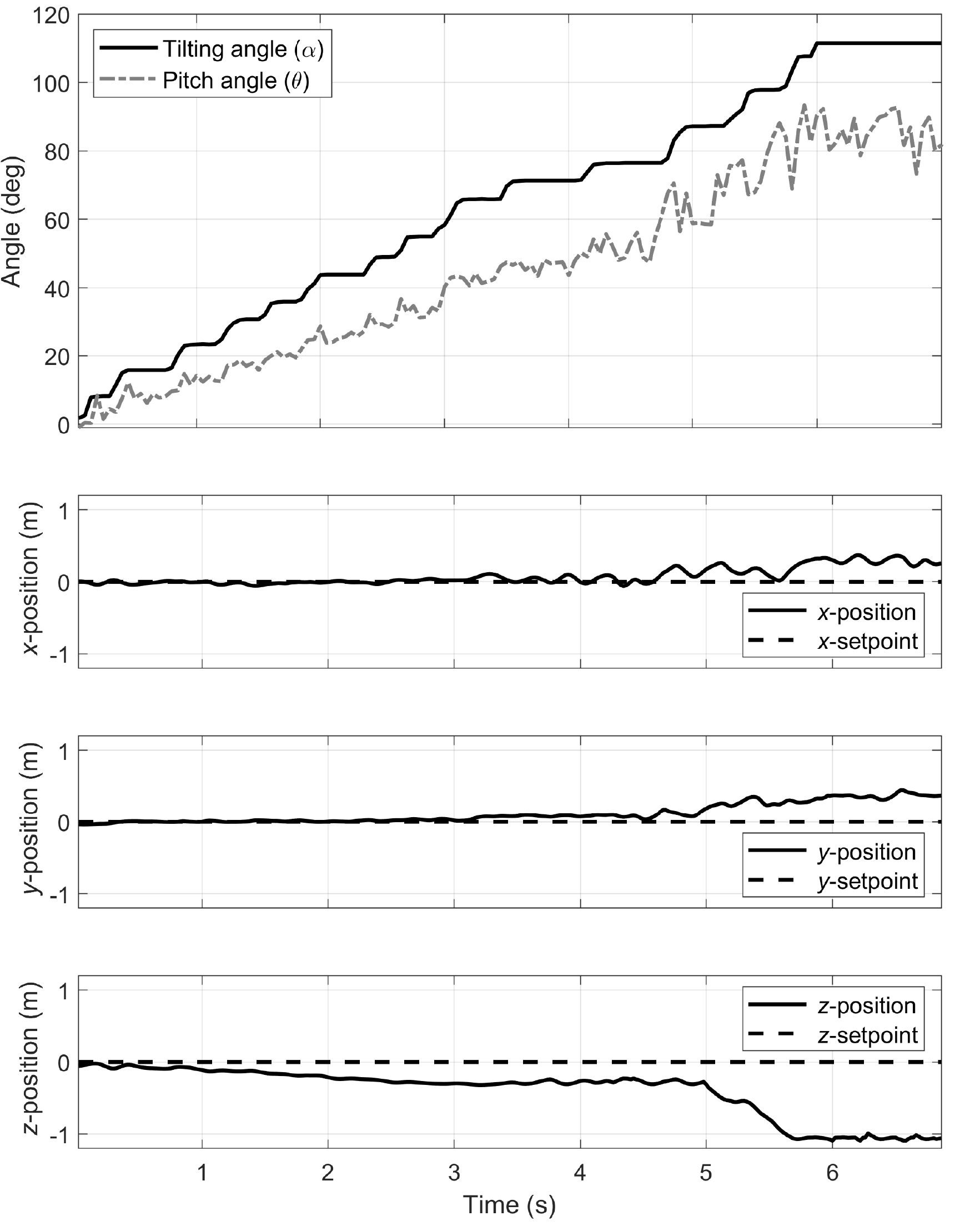}
        \caption{Actual flight results of hovering without learning real-world data. The pitch angle $\theta$ according to the given tiling angle $\alpha$ and corresponding position at that time.}
        \label{exp1}
        \vspace{-0.5cm}
    \end{figure}
To check the degree of sim-to-real gap, the same task as in Fig.~\ref{sim_log} has been tested. The results  are  shown  in  Fig.~\ref{exp1}. The overall trend of the results is similar to that of the simulation in Fig.~\ref{sim_log}. However, there is a significant difference between the simulation and the actual flight owing to the sim-to-real gap caused by nonlinear effects that have not been modeled in the simulation. These nonlinearities have occurred from the interference between propellers by the tilting rotor, the aerodynamics of coaxial propeller, and the difference between modeled center of mass and the actual center of mass.

\subsection{With \textcolor{black}{Real-world Data}}
\begin{figure}[t] 
    \centering
        \includegraphics[scale=0.41]{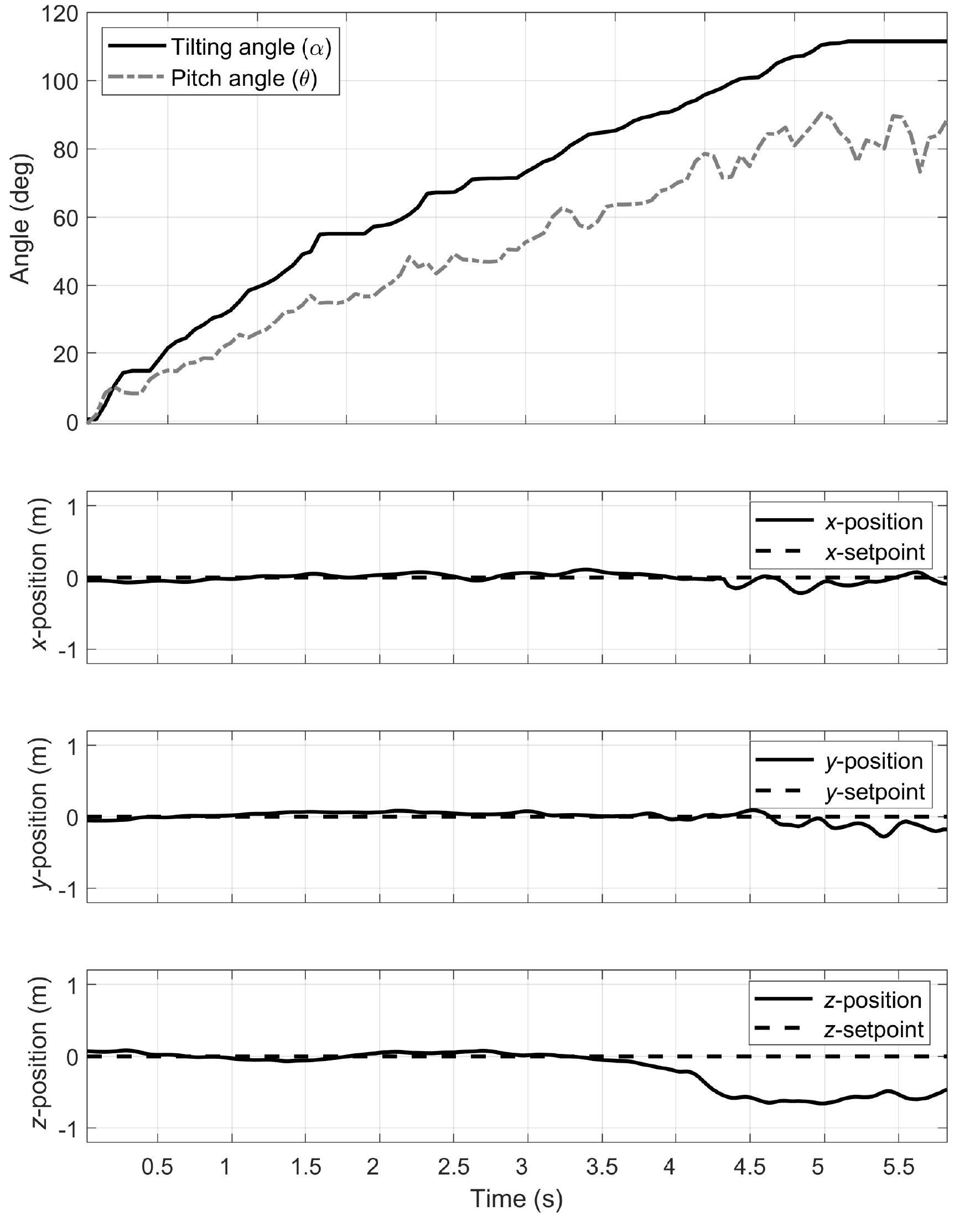}
        \caption{Actual flight results of hovering after learning real-world data. The pitch angle $\theta$ according to the given tiling angle $\alpha$ and corresponding position at that time.}
        \label{fig11}
    \end{figure}
    
    
\begin{figure}
    \centering
    \subfigure[]{\includegraphics[width=0.8in]{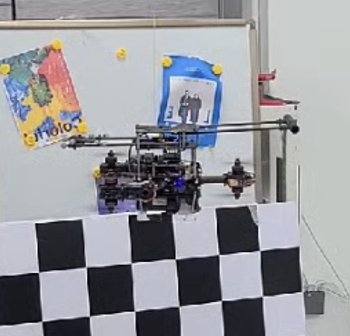}}
    \subfigure[]{\includegraphics[width=0.8in]{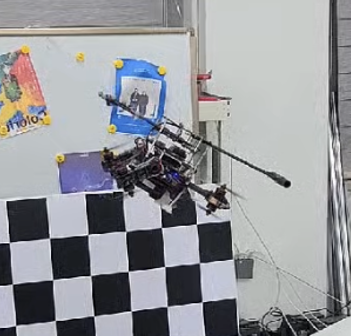}} 
    \subfigure[]{\includegraphics[width=0.8in]{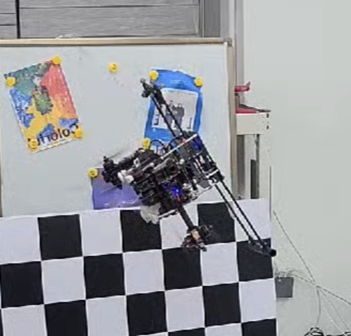}}
    \subfigure[]{\includegraphics[width=0.8in]{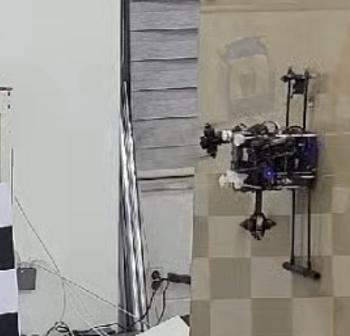}}
        \vspace{-0.2cm}
    \caption{Perching process of the platform from hovering state. (a) Beginning, (b) After 1.5 s, (c) After 3.5 s, (d) After perching.}
    \label{fig12}
    \vspace{-0.5cm}
\end{figure}
    
The policy network was additionally trained by the actual flight data. There is a difficulty in obtaining actual flight data for a multirotor sufficiently due to flight time limitations. We checked whether we could learn enough to achieve acceptable performance for perching by learning with limited real-world data. To acquire the data for the training, flights with the platform were repeated. Each flight was about 3 minutes long, and 30 times of flights have been conducted. Because the state vector is updated at a 200Hz rate, a total of about 1 million actual flight data has been obtained. Afterward, 2 million data were achieved with data augmentation using symmetry. The final experimental results after learning these data are shown in Fig. \ref{fig11}. The platform hovered and maintained its position more stably and accurately up to 90\textdegree\space compared to the one without learning real-world data. The result was sufficient to tilt in the air and perch to the wall, as shown in Fig. \ref{fig12}. With these results, we could conclude that the nonlinear effects that were not modeled in the simulation were eventually learned through the real-world data. In this research, we focused on demonstrating that the tilting multirotor can be controlled through reinforcement learning. 

\section{Conclusion}

This study proposed a novel control strategy for controlling a tilting multirotor with reinforcement learning. Controlling an actual system with reinforcement learning has the advantage of not needing to know the actual model accurately; however, it may not be appropriately controlled due to a sim-to-real gap. In this paper, the gap was overcome by learning real experiences from the actual flight. Additionally, it is challenging and requires more data to learn the tilting multirotor dynamics because it is more complicated than the existing quadrotor. As described in Section \uppercase\expandafter{\romannumeral3}.\textit{B}. and \uppercase\expandafter{\romannumeral3}.\textit{C}., novel methods to learn faster and augment the data have been proposed. The results of the control using these methods are presented in Section V. In this study, only the task involving perching on the wall by tilting in the air was demonstrated; however, an additional controller suitable for the wall is required for applications on an actual wall. Future work will enable perching, detaching, and even wall interaction with a single controller without an additional one.





\end{document}